\documentclass[runningheads]{llncs}

\usepackage{eccv}

\usepackage{eccvabbrv}

\usepackage{graphicx}
\usepackage{booktabs}
\usepackage{tcolorbox}
\usepackage{tikz}
\usepackage{pgfplots}
\usepackage{bm}
\usepackage{url}

\usepgfplotslibrary{groupplots}
\pgfplotsset{compat=1.18}
\usepackage{xcolor} 
\usepackage[accsupp]{axessibility}  
\usepackage{enumitem}
\usepackage{soul}

\usepackage[breaklinks,colorlinks,citecolor=eccvblue]{hyperref}

\usepackage{orcidlink}

\sethlcolor{red!20} 
\usepackage[table]{xcolor}

\title{CAViAR: A Causal Video Dataset for Fine-Grained Accident Reasoning in Real-World Scenarios}
\titlerunning{CAViAR}

\author{Sparsh Garg\thanks{Corresponding author. email: sparsh \textit{at} nec-labs \textit{dot} com} \and
Yi-Wen Chen \and
Vijay Kumar B G \and
Abhishek Aich}

\authorrunning{S. Garg et al.}

\institute{NEC Laboratories, America, USA}
\begin{document}
\maketitle
\begin{abstract}

While modern autonomous driving systems excel at perception tasks such as object detection and trajectory prediction, they lack the high-level causal reasoning required to interpret traffic accidents. In particular, determining responsibility, such as identifying who is at fault and which traffic rule was violated, remains largely unexplored in current benchmarks. To this end, we introduce \textbf{CAViAR} (Causal Accident Video and Incident Analysis Repository), a human-annotated dashcam benchmark comprising 2,249 real-world accident videos collected from CarCrashDataset (CCD) and Nexar. Each video is annotated with structured labels spanning environmental conditions, accident type, causal explanation, apparent At-Fault Agent, affected agent, and apparent rule-violation category. We benchmark state-of-the-art vision-language models (VLMs), including Cosmos-Reason2, Qwen3-VL, and InternVL3. Once class imbalance is accounted for with majority/random baselines and balanced metrics, perceptual competence is uneven--lighting is nearly solved, whereas weather and road-condition accuracy fall at or below the majority-class baseline---and all models degrade sharply on accident type and responsibility reasoning. Overall, \textbf{CAViAR} exposes a practical Perception--Reasoning Gap: current VLMs may recognize salient context, but do not reliably map visible agent actions to annotated rule-relevant responsibility categories in safety-critical driving scenarios.
Code, annotation schema, prompts, and evaluation scripts are available at: \url{https://github.com/nec-labs-ma/CAViAR}.
\end{abstract}    
\section{Introduction}

Traffic accidents remain a major global safety concern, motivating extensive research in collision detection, anomaly prediction, and trajectory forecasting. While modern vision systems excel at identifying objects and predicting motion, they largely operate at the level of perception. In contrast, real-world deployment---whether in autonomous driving, surveillance, or forensic analysis---requires \textit{causal and responsibility reasoning}: understanding why an accident occurred and which observable agent appears responsible.

Existing accident benchmarks largely emphasize detection or coarse categorization, and do not explicitly evaluate rule-grounded responsibility reasoning such as apparent at-fault agent, and affected agent identification. To address this limitation, we introduce \textbf{CAViAR} (Causal Accident Video and Incident Analysis Repository), a benchmark of \textbf{2,249 real-world dashcam videos} built on top of CarCrashDataset (CCD)~\cite{BaoMM2020} (1,500 videos) and Nexar~\cite{moura2025nexar} (749 videos after filtering). We reuse these established sources to ensure stable provenance and reproducible comparisons, while contributing new responsibility-oriented supervision that is absent from prior benchmarks. Each video in CAViAR is annotated with structured supervision spanning (i) environment (weather, lighting, road condition), (ii) accident analysis (accident type and causal explanation), and (iii) responsibility attribution (at-fault agent, affected agent, and apparent traffic rule violation). This enables evaluation tasks that go beyond perception:
\begin{itemize}
    \item \textbf{Perception:} What are the environmental conditions and accident type?
    \item \textbf{Causality:} What sequence of actions led to the incident?
    \item \textbf{Responsibility:} Who is at fault and which traffic rule was violated?
\end{itemize}

By integrating perception, causality, and responsibility into a unified benchmark, CAViAR enables systematic evaluation of \textit{vision-language models (VLMs)} on safety-critical reasoning. In particular, it allows us to quantify a \textit{Perception--Reasoning Gap}: models can often recognize contextual cues (e.g., weather and lighting) yet struggle to apply explicit traffic rules for responsibility attribution. Table~\ref{tab:dataset_comparison} positions CAViAR against prior accident datasets, highlighting that none provide structured fault and rule-violation supervision. Overall, CAViAR bridges raw visual evidence and responsibility-oriented annotations, providing a benchmark for advancing rule-grounded multimodal reasoning in autonomous safety.

\section{Related Work}

\noindent\textbf{Accident datasets.}
Traffic accident understanding has been studied through real-world and synthetic benchmarks. Dashcam datasets such as ROL~\cite{karim2023rol}, A3D~\cite{yao2019unsupervised}, DoTA~\cite{yao2022dota}, and DADA-2000~\cite{fang2019dada} focus mainly on detection, anticipation, or temporal localization, while synthetic datasets such as DeepAccident~\cite{wang2024deepaccident} and CTAD~\cite{ctad} support controlled evaluation of rare safety-critical events. These datasets are valuable for recognizing \textit{what} happened, but provide limited supervision for explaining \textit{why} it happened or \textit{who} is responsible.
The Car Crash Dataset (CCD)~\cite{BaoMM2020} contains 1,500 real-world dashcam clips with binary normal/anomaly annotations and temporal labels; it supplies collision and near-miss clips but no responsibility or rule-violation labels.
Nexar~\cite{moura2025nexar} is a large-scale multi-city dashcam corpus capturing naturalistic driving including crashes; it provides geo-diversity and verified collision clips but likewise no structured QA or causal annotations. CAViAR builds on both as source footage (CCD for training, Nexar for testing) while contributing a new responsibility-reasoning annotation layer absent from either.

\noindent\textbf{VideoQA for responsibility reasoning.}
Traffic-oriented VideoQA datasets, including CTA~\cite{you2020CTA}, SUTD-TrafficQA~\cite{xu2021sutdtrafficqa}, TUMTraffic-VideoQA~\cite{zhou2025tumtraffic}, MM-AU~\cite{Fang_2024_CVPR}, and VRU-Accident~\cite{kim2025vru}, evaluate multimodal understanding of road scenes. However, existing benchmarks do not provide unified supervision for fault identification, victim identification, and violated traffic rules. CAViAR fills this gap by explicitly targeting rule-grounded responsibility reasoning in accident videos.

\begin{table*}[t]
\centering
\caption{Comparison of accident video datasets. CAViAR uniquely provides structured supervision for \emph{fault attribution} and \emph{traffic rule violation identification}.}
\label{tab:dataset_comparison}
\small
\setlength{\tabcolsep}{4pt}
\resizebox{\textwidth}{!}{
\begin{tabular}{lcccccccc}
\toprule
\textbf{Dataset} & \textbf{View} & \textbf{\#Clips} & \textbf{VQA} & \textbf{Causal} & \textbf{Dense Cap.} & \textbf{At-Fault} & \textbf{Rule Viol.} & \textbf{Real/Synth.}\\
\midrule
CCD~\cite{BaoMM2020} & Dashcam & 1,500 & N & N & N & N & N & Real \\
Nexar~\cite{moura2025nexar} & Dashcam & 5M+ & N & N & N & N & N & Real \\
ROL~\cite{karim2023rol} & Dashcam & 1,000 & N & N & N & N & N & Real \\
DeepAccident~\cite{wang2024deepaccident} & Dashcam & -- & N & N & N & N & N & Synthetic \\
CTA~\cite{you2020CTA} & Dashcam & 1,935 & Y & N & N & N & N & Real \\
CTAD~\cite{ctad} & Surveillance & 1,100 & N & N & N & N & N & Synthetic \\
SUTD-TrafficQA~\cite{xu2021sutdtrafficqa} & Surveillance & 10,080 & Y & N & N & N & N & Real \\
TUMTraffic-VideoQA~\cite{zhou2025tumtraffic} & Surveillance & 1,000 & Y & Y & Y & N & N & Real \\
TUMTraf-A~\cite{zimmer2025towards} & Surveillance & 48 & N & N & N & N & N & Real \\
A3D~\cite{yao2019unsupervised} & Dashcam & 3,757 & N & N & N & N & N & Real \\
DoTA~\cite{yao2022dota} & Dashcam & 5,586 & N & N & N & N & N & Real \\
DADA-2000~\cite{fang2019dada} & Dashcam & 2,000 & N & Y & N & N & N & Real \\
MM-AU~\cite{Fang_2024_CVPR} & Dashcam & 11,727 & Y & Y & N & N & N & Real \\
VRU-Accident~\cite{kim2025vru} & Dashcam & 1,000 & Y & Y & Y & N & N & Real \\
\midrule
\textbf{CAViAR (Ours)} & Dashcam & \textbf{2,249} & Y & Y & Y & \textbf{Y} & \textbf{Y} & Real \\
\bottomrule
\end{tabular}
}
\vspace{-0.5em}
\end{table*}
\section{The CAViAR Dataset}
\label{sec:dataset}

CAViAR (\emph{Causal Accident Video and Incident Analysis Repository}) is a
real-world dashcam benchmark of \textbf{2{,}249} videos with
\textbf{20{,}108} question--answer (QA) pairs. The benchmark contains
nine QA prompts grouped into eight task families. Unlike prior accident
corpora that stop at detection, anticipation, or generic VideoQA, CAViAR
adds an explicit \emph{video-grounded responsibility reasoning} layer:
apparent at-fault agent, affected agent, and apparent rule-violation
category. These labels are annotations of observable responsibility cues
from video evidence and are not legal determinations of liability.

\begin{figure}[t]
\centering
\includegraphics[width=\linewidth]{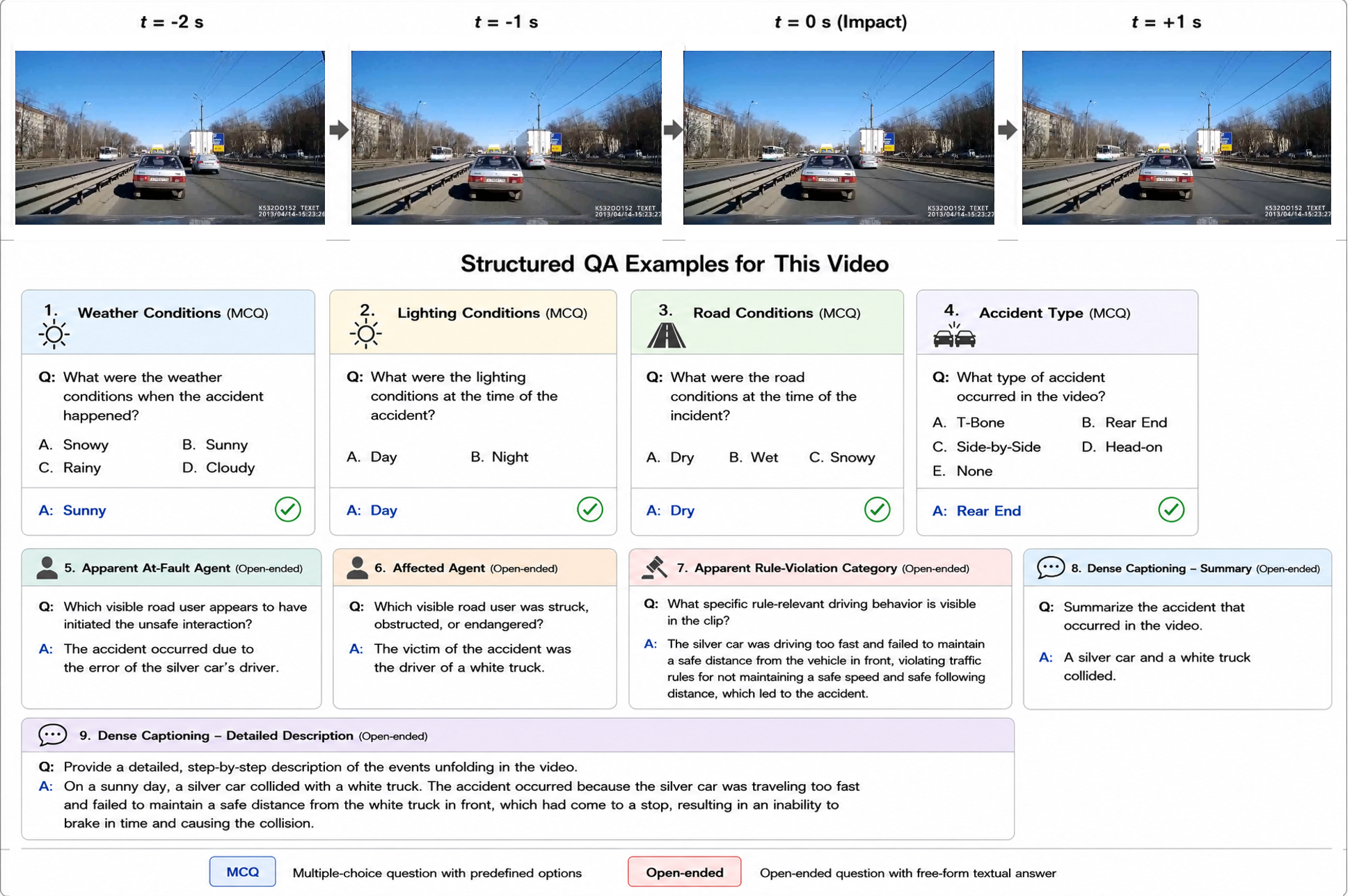}
\caption{Example CAViAR video with its nine QA prompts grouped into eight
task families, spanning dense captioning, environmental conditions,
accident type, and video-grounded responsibility annotations
(apparent at-fault agent, affected agent, and apparent rule-violation
category).}
\label{fig:qa_examples}
\end{figure}

We describe the source data and rationale (Sec.~\ref{sec:src}),
collection and filtering (Sec.~\ref{sec:filter}), annotation pipeline
(Sec.~\ref{sec:annot}), QA structure
(Sec.~\ref{sec:tasks}), and statistics, splits and task justification
(Sec.~\ref{sec:stats}). Figure~\ref{fig:qa_examples} shows a
representative video with its complete multi-task annotation.

\subsection{Source Data and Rationale}
\label{sec:src}

CAViAR is built on two public dashcam corpora, the Car Crash Dataset
(CCD)~\cite{BaoMM2020} and Nexar~\cite{moura2025nexar}, which provide
verified collision and near-collision clips and avoid the cost and ethical
burden of mining rare crash events from raw footage. Their complementary
fleet, geographic, and camera distributions enable a leakage-free split:
CCD for training and Nexar for testing, with no shared video, scene, or
device. Our contribution is not raw video collection but a new
video-grounded responsibility-reasoning annotation layer over trusted
footage; we comply with the original licenses and release annotations,
splits, prompts, and frame indices rather than redistributing restricted
video content.

\subsection{Collection and Filtering}
\label{sec:filter}
Both corpora are filtered to retain clips with a collision or a
safety-critical near-miss, yielding \textbf{1{,}500} CCD and \textbf{749}
Nexar clips (\textbf{2{,}249} total). We verify that each clip shows
sufficient pre-incident context, the critical interaction window, and the
aftermath when available, and that the involved agents are visually
distinguishable so responsibility fields are answerable from visible
evidence. Selection criteria and final clip ID lists will be released.

\subsection{Annotation Pipeline}
\label{sec:annot}

\textbf{Annotator roles.}
CAViAR is annotated by a four-person team with traffic-safety familiarity.
Two primary annotators produce the initial labels and free-form answers.
Two additional reviewers perform continuous quality-control checks over
the annotated clips, focusing on cross-field consistency, visual support,
and ambiguous responsibility cases. Thus, CAViAR follows a
primary-annotation plus review protocol rather than a fully independent
multi-rater protocol for every clip.

\textbf{Stage 1 -- Primary human annotation.}
For each clip, the primary annotators identify the involved agents,
summarize the event, assign MCQ labels for environmental conditions and
accident type, and write open-ended answers for the apparent at-fault
agent, affected agent, and apparent rule-violation category. The
\emph{apparent at-fault agent} is the road user whose visible action most
directly initiated the unsafe interaction. The \emph{affected agent} is
the road user visibly struck, obstructed, or endangered by that action.
The \emph{apparent rule-violation category} describes the rule-relevant
driving behavior visible in the clip.

\textbf{Jurisdiction-agnostic rule ontology.}
Rather than jurisdiction-specific legal statutes, annotators use a
jurisdiction-agnostic ontology of eleven families of visually observable
rule-relevant driving behaviors, each defined by characteristic visual
preconditions and typical evidence; the families and the deterministic
mapping protocol are detailed in Sec.~\ref{sec:rulediv}.

\textbf{Stage 2 -- Quality-control review and adjudication.}
The two reviewers check that the dense caption, at-fault agent, affected
agent, and rule-violation category refer to the same visible agents and
describe a compatible causal sequence (e.g., a ``failed to maintain safe
following distance'' violation must name the follower as at-fault and the
leader as affected). Inconsistent, unsupported, or unclear annotations are
returned for revision and resolved through team discussion.

\textbf{Ambiguity handling.}
Not every accident is decidable from a dashcam view alone. Clips whose
responsible party cannot be established from visible evidence (occluded
interactions, off-screen initiators, symmetric maneuvers, or insufficient
context) are flagged and excluded from responsibility-oriented evaluation
rather than forced into a single-agent label; where evidence supports
shared responsibility, the annotation records this explicitly.

\textbf{Stage 3 -- LLM language refinement.}
Raw human-written annotations can be terse or grammatically inconsistent.
We use GPT-4~\cite{achiam2023gpt} only to normalize grammar and phrasing,
explicitly instructing it not to add, infer, or alter facts. The prompt is:
\begin{quote}\itshape\small
``Rewrite the following text into fluent English. Do not introduce any
vehicle, action, or cause not present in the original text. Preserve all
entities and the stated at-fault agent, affected agent, and rule-violation
category exactly.''
\end{quote}
The human-written labels remain the source of truth. Where possible, we
release both the raw human annotation and the GPT-refined text fields to
make the refinement step auditable.

\subsection{QA Structure}
\label{sec:tasks}

Each video is mapped to a fixed set of QA records stored in a common schema:
\begin{verbatim}
{ "benchmark": <task>, "question": <text>, "answer": <ref>,
  "choices": [...], "correct_answer": <opt>, "correct_index": <int> }
\end{verbatim}
MCQ tasks include a closed option set and correct index, while open-ended
tasks store the human-written reference answer. CAViAR contains nine QA
prompts grouped into eight task families: dense-captioning prompts,
four MCQ prompts, and three video-grounded responsibility prompts. The
apparent rule-violation category also serves as the rule-relevant
accident-reason annotation; we therefore do not count ``Accident Reason''
as a separate task. Table~\ref{tab:caviar_taxonomy} provides the canonical
task taxonomy.

\subsection{Statistics, Splits, and Task Justification}
\label{sec:stats}

Over the leakage-free CCD/Nexar split (Sec.~\ref{sec:src}), CAViAR contains
\textbf{8{,}996} MCQ questions and \textbf{11{,}112} open-ended questions.
The canonical task taxonomy, split counts, missing labels, and evaluation
metrics are reported in Table~\ref{tab:caviar_taxonomy}.

\begin{table*}[t]
\centering
\caption{Canonical CAViAR task taxonomy and statistics. CAViAR contains nine QA prompts grouped into eight task families.}
\label{tab:caviar_taxonomy}
\scriptsize
\setlength{\tabcolsep}{3pt}
\resizebox{\textwidth}{!}{
\begin{tabular}{llp{4.1cm}lcccc}
\toprule
\textbf{ID} & \textbf{Task family} & \textbf{Question template / label space} & \textbf{Type} & \textbf{Train} & \textbf{Test} & \textbf{Missing} & \textbf{Metric} \\
\midrule
T1a/T1b & Dense Captioning &
Detailed event description and brief accident summary. &
Open & 2{,}991 & 1{,}475 & 32 & BERTScore-F1 \\

T2 & Weather &
What is the weather condition? \{sunny, rainy, cloudy, snowy\}. &
MCQ & 1{,}500 & 749 & 0 & Acc. \\

T3 & Lighting &
Is the video taken during day or night? \{day, night\}. &
MCQ & 1{,}500 & 749 & 0 & Acc. \\

T4 & Road Condition &
What is the road condition? \{dry, wet, snowy\}. &
MCQ & 1{,}500 & 749 & 0 & Acc. \\

T5 & Accident Type &
What type of accident occurred? \{rear-end, T-bone, side-by-side, head-on, none\}. &
MCQ & 1{,}500 & 749 & 0 & Acc. \\

T6 & Apparent At-Fault Agent &
Which visible road user appears to have initiated the unsafe interaction? &
Open & 1{,}483 & 724 & 42 & Judge \\

T7 & Affected Agent &
Which visible road user was struck, obstructed, or endangered? &
Open & 1{,}472 & 724 & 53 & Judge \\

T8 & Apparent Rule-Violation Category &
Which rule-relevant behavior is visible? Uses the jurisdiction-agnostic ontology in Sec.~\ref{sec:annot}. &
Open & 1{,}499 & 744 & 6 & Judge \\
\midrule
\textbf{Total} & -- & -- & -- & \textbf{13{,}445} & \textbf{6{,}663} & \textbf{133} & -- \\
\bottomrule
\end{tabular}
}
\end{table*}

The maximum possible number of QA pairs is
$2{,}249 \times 9 = 20{,}241$; CAViAR contains \textbf{20{,}108} because
\textbf{133} fields are unavailable (per-task counts in Table~\ref{tab:caviar_taxonomy}).
Missing responsibility annotations correspond almost entirely to near-miss
(\texttt{none}) clips, where no collision occurs and thus no responsibility
field is defined; these fields are left unlabeled rather than forced,
consistent with our ambiguity-handling protocol (Sec.~\ref{sec:dataset}).

\noindent\textbf{Why include environmental tasks?}
Weather, lighting, and road condition are retained as contextual
conditioning variables, not as the central challenge. First,
responsibility judgments may depend on them, such as following distance
on a wet road or visibility at night. Second, they act as a difficulty
calibration floor: high accuracy on these easier visual cues alongside low
performance on responsibility tasks helps isolate the reasoning gap
CAViAR targets. We therefore report them separately and do not fold them
into the headline responsibility score.

\subsection{Comparison with Existing Datasets}
\label{sec:compare}

Table~\ref{tab:dataset_comparison} summarizes key differences between
CAViAR and existing accident video datasets. While prior benchmarks cover
detection, temporal localization, accident anticipation, or traffic
VideoQA, none provide unified supervision for apparent at-fault agent,
affected agent, and apparent rule-violation category. CAViAR integrates
these video-grounded responsibility annotations alongside perception and
description tasks to support unified evaluation of accident understanding
and rule-relevant multimodal reasoning.

\section{Experiments}
\label{sec:exp}

The goal of this section is to \emph{validate CAViAR as a benchmark}: to show that it cleanly separates perception from responsibility reasoning and exposes failure modes that current VLMs share, rather than to propose a new model.

\subsection{Experimental Setup}

\noindent\textbf{Dataset split.}
Models are trained on CCD~\cite{BaoMM2020} (1,500 videos) and evaluated on held-out Nexar~\cite{moura2025nexar} (749 videos after filtering). Because the train and test splits come from disjoint sources with no shared video, scene, or device, the split is leakage-free by construction.

\noindent\textbf{Models and rationale.}
We benchmark three open-weight video VLM families at two scales each (2B/8B): Cosmos-Reason2~\cite{cosmos_reason2_2026}, Qwen3-VL~\cite{bai2025qwen3}, and InternVL3~\cite{zhu2025internvl3}. They are openly released (reproducible fine-tuning), span complementary designs (reasoning-tuned, generalist, and strong open VLM), and are widely adopted; the two scales per family separate model scale from domain adaptation. We report zero-shot (\textit{Base}) and LoRA supervised fine-tuning (\textit{Fine-tuned}) on CCD, freezing the vision encoder and projector and adapting only the language-model attention and MLP projections. We use a fixed training recipe with no Nexar examples used for tuning or checkpoint selection, and greedy decoding with fixed seeds. To expose class-prior or option-prior effects, we additionally report random/majority baselines, balanced accuracy, and macro-F1.

\noindent\textbf{Metrics.}
We report \textbf{MCQ accuracy} for Weather, Lighting, Road Conditions, and Accident Type; \textbf{BERTScore-F1}~\cite{zhang2019bertscore} for open-ended generation; and \textbf{LLM-as-Judge} (GPT-4o~\cite{hurst2024gpt}, 0--5) for responsibility reasoning. We report these three jointly because no single open-ended metric fully captures the rule-grounded content of responsibility answers (Sec.~\ref{sec:limitations}).

\noindent\textbf{MCQ evaluation and invalid outputs.}
For MCQs, we extract the first valid option letter from the model output using a regex over \{A,B,C,D,E\}. If no valid option is found (e.g., a small model that emits free text instead of a letter), the prediction is counted as incorrect. This treats malformed answers conservatively and ensures consistent evaluation across model sizes.

\noindent\textbf{Frame sampling and decoding.}
At 16 FPS we sample frames uniformly for feasibility: InternVL3 uses segment-midpoint sampling capped at 64 frames, while Qwen3-VL and Cosmos-Reason2 use the processor's internal extraction (an \texttt{fps} parameter). Decoding is greedy (\texttt{do\_sample=False}); \texttt{max\_new\_tokens} is $512$ for InternVL3 and $4096$ for Qwen3-VL/Cosmos-Reason2, which emit a \texttt{<think>} trace that is stripped before judging, so the differing budgets do not affect scores.

\noindent\textbf{Fine-tuning.}
We apply LoRA SFT~\cite{hu2022lora} on CCD (no RL or preference optimization), freezing the vision encoder and projector and inserting adapters into the language-model attention/MLP projections (\texttt{\{q,k,v,o\}\_proj}, \texttt{\{gate,up,down\}\_proj}): rank 64, $\alpha{=}128$, dropout 0.05; AdamW, \texttt{bf16}, gradient checkpointing; learning rate 2e-5 (2B)/1e-5 (8B), cosine schedule, warmup 0.03 (InternVL3)/0.10 (Qwen3-VL/Cosmos), 1 epoch (InternVL3) and 3 epochs(Qwen3 and Cosmos), global batch 128/16--32 via gradient accumulation, seed 42 (InternVL3 with DeepSpeed ZeRO-1; Qwen3-VL/Cosmos with TRL\,+\,ZeRO-2). We use a fixed recipe with \emph{no} held-out validation split or early stopping (Nexar is used only for final evaluation), so the limited 8B gains should be read with overfitting/under-tuning as a live risk. We report one run per model and quantify uncertainty with non-parametric bootstrap $95\%$ CIs over test examples ($10{,}000$ resamples); half-widths ($\approx\!\pm1.7$ MCQ points, $\approx\!\pm0.07$ judge) exceed all three 8B MCQ deltas, the basis for calling those gains marginal.

\noindent\textbf{Prompts and judge.}
All models receive identical prompts (aside from model-specific chat templates): a task context sentence, the question, and a format instruction. MCQ options are listed \texttt{A./B./\dots} in a \emph{fixed canonical order} (not randomized per item); we parse the predicted letter back to its option text and report balanced accuracy and macro-F1 to penalize collapse onto a frequent option. Open-ended prompts ask the model to identify the apparent at-fault agent, the affected agent, or the annotated rule-relevant behavior category and its associated visible road user, with no extra system prompt. The GPT-4o judge~\cite{hurst2024gpt} returns \texttt{Score:\ <0--5>} (5 = correct agent(s) and reasoning; 3 = correct agent but weak reasoning; 0 = irrelevant).

\begin{table*}[t]
\centering
\caption{Main results on Nexar at 16 FPS. We report MCQ accuracy (\%), BERTScore-F1 (\%), and LLM-as-Judge (0--5) for Base vs. fine-tuned (FT) models.}
\label{tab:main_results}
\small
\setlength{\tabcolsep}{3pt}
\resizebox{\textwidth}{!}{
\begin{tabular}{lccccccccc}
\toprule
& \multicolumn{3}{c}{\textbf{MCQ Acc.}} 
& \multicolumn{3}{c}{\textbf{BERTScore-F1}} 
& \multicolumn{3}{c}{\textbf{Judge}} \\
\cmidrule(lr){2-4}\cmidrule(lr){5-7}\cmidrule(lr){8-10}
\textbf{Model} 
& Base & FT & $\Delta$ 
& Base & FT & $\Delta$ 
& Base & FT & $\Delta$ \\
\midrule
Cosmos2-2B   & 66.66 & \textbf{74.80} & +8.14  & 29.48 & \textbf{47.73} & +18.25 & 1.396 & \textbf{1.814} & +0.418 \\
Cosmos2-8B   & 65.29 & 65.79 & +0.50  & 27.04 & 27.66 & +0.62  & 1.127 & 1.382 & +0.255 \\
Qwen3-2B     & 55.67 & 67.16 & +11.49 & 20.07 & 45.36 & +25.29 & 1.280 & 1.763 & +0.483 \\
Qwen3-8B     & 61.42 & 62.32 & +0.90  & 12.85 & 19.71 & +6.86  & 1.150 & 1.397 & +0.247 \\
InternVL3-2B & 61.95 & 62.68 & +0.73  & 29.59 & 33.53 & +3.94  & 1.203 & 1.318 & +0.115 \\
InternVL3-8B & \textbf{69.06} & 68.83 & $-$0.23 & \textbf{35.12} & 35.72 & +0.60 & \textbf{1.527} & 1.549 & +0.022 \\
\bottomrule
\end{tabular}
}
\vspace{-0.5em}
\end{table*}

\begin{figure}[t]
\centering
\includegraphics[width=\linewidth]{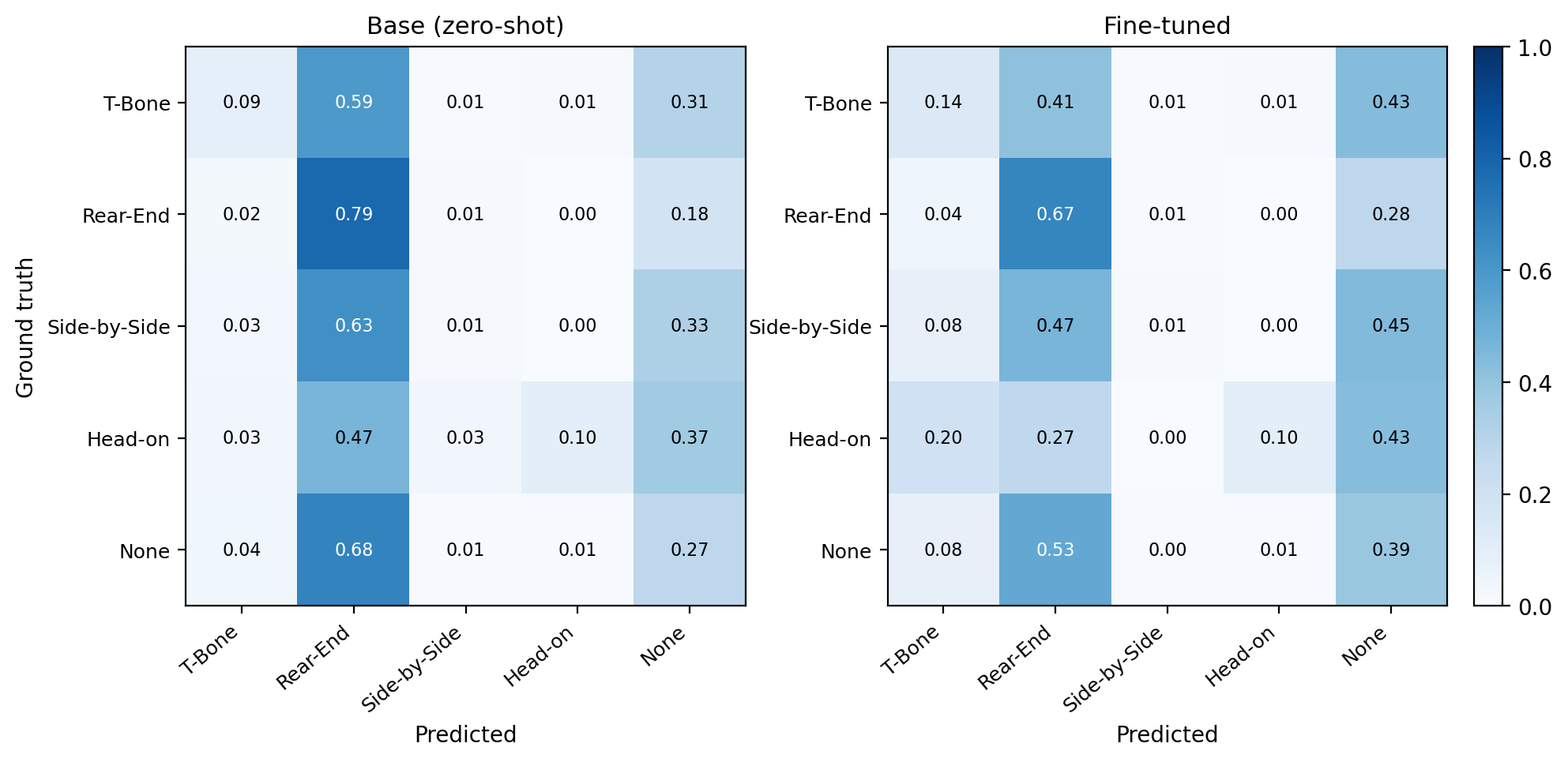}
\caption{Row-normalized accident-type confusion matrices, aggregated over all six models (base vs.\ fine-tuned) at 16 FPS ($N{=}4{,}494$ predictions each). Both regimes collapse onto \emph{Rear-End} and rarely recover \emph{Side-by-Side} or \emph{Head-on} (quantitative details in text).}
\label{fig:confusion}
\end{figure}

\subsection{Quantitative Results}

\paragraph{Per-task MCQ diagnostics.}
Because the MCQ tasks differ substantially in difficulty, we report each
separately with random and majority-class baselines
(Table~\ref{tab:mcq_diagnostics}) rather than using aggregate MCQ accuracy
as evidence of uniformly strong perception.

\begin{table*}[t]
\centering
\begin{minipage}[t]{0.45\textwidth}
\centering
\caption{LLM-as-Judge scores (0--5) by responsibility subtask at 16 FPS. Violation identification is the weakest subtask for every model.}
\label{tab:judge_breakdown}
\resizebox{\linewidth}{!}{%
\begin{tabular}{lccc}
\toprule
\textbf{Model} & \textbf{At-Fault} & \textbf{Affected} & \textbf{Rule Viol.} \\
\midrule
InternVL3-2B & 1.39 & 1.82 & 0.42 \\
InternVL3-2B (FT) & 1.62 & 1.83 & 0.52 \\
InternVL3-8B & 2.00 & 1.93 & 0.67 \\
InternVL3-8B (FT) & 1.94 & 1.98 & 0.75 \\
Qwen3-VL-2B & 1.58 & 1.79 & 0.49 \\
Qwen3-VL-2B (FT) & \textbf{2.27} & 2.26 & 0.79 \\
Qwen3-VL-8B & 1.16 & 1.64 & 0.67 \\
Qwen3-VL-8B (FT) & 1.38 & 2.20 & 0.64 \\
Cosmos2-2B & 1.98 & 1.64 & 0.59 \\
Cosmos2-2B (FT) & 1.95 & \textbf{2.70} & \textbf{0.82} \\
Cosmos2-8B & 1.42 & 1.37 & 0.60 \\
Cosmos2-8B (FT) & 1.55 & 1.85 & 0.76 \\
\bottomrule
\end{tabular}}
\end{minipage}\hfill
\begin{minipage}[t]{0.53\textwidth}
\centering
\caption{Per-task MCQ diagnostics on Nexar, averaged over the six models (Base/FT). Acc, balanced accuracy (Bal-Acc), and macro-F1 are over classes present in the test split; majority/random baselines expose class-prior effects hidden by aggregate accuracy.}
\label{tab:mcq_diagnostics}
\resizebox{\linewidth}{!}{%
\begin{tabular}{lccccc}
\toprule
\textbf{Task} & \textbf{Random} & \textbf{Majority} & \textbf{Acc} & \textbf{Bal-Acc} & \textbf{Macro-F1} \\
 & & & {\scriptsize Base/FT} & {\scriptsize Base/FT} & {\scriptsize Base/FT} \\
\midrule
Weather       & 25.0 & 73.0 & 58.4\,/\,62.6 & 64.0\,/\,62.9 & 49.3\,/\,50.3 \\
Lighting      & 50.0 & 62.6 & 98.6\,/\,98.7 & 98.7\,/\,98.7 & 98.5\,/\,98.6 \\
Road Cond.    & 33.3 & 89.3 & 65.3\,/\,74.0 & 80.9\,/\,83.8 & 59.8\,/\,65.8 \\
Accident Type & 20.0 & 37.7 & 33.1\,/\,35.4 & 25.1\,/\,26.1 & 18.6\,/\,21.1 \\
\bottomrule
\end{tabular}}
\end{minipage}
\end{table*}

Table~\ref{tab:mcq_diagnostics} shows that Lighting is nearly solved across
models, but Weather accuracy remains below the majority-class baseline and Road
Condition accuracy is also below the majority baseline despite higher balanced
accuracy. Accident Type is the most difficult MCQ task: raw accuracy is near the
majority baseline, but balanced accuracy and macro-F1 remain low. We therefore
interpret aggregate MCQ results cautiously.

\paragraph{Overall performance and scaling.}
Table~\ref{tab:main_results} shows a consistent Perception--Reasoning Gap: MCQ accuracy is comparatively high, while responsibility reasoning remains low in absolute terms. Fine-tuning substantially improves the 2B models, whereas 8B gains are limited and consistent across families (e.g., InternVL3-8B MCQ $\Delta{=}{-}0.23$, Qwen3-8B $\Delta{=}{+}0.90$), indicating that scaling alone is insufficient for responsibility reasoning (causes analyzed in Sec.~\ref{sec:discussion}). Within responsibility, violation identification is consistently the weakest subtask (Table~\ref{tab:judge_breakdown}), typically around $0.8/5.0$ even after fine-tuning.

\paragraph{Interpreting the judge scores.}
Responsibility scores use the $0$--$5$ judge rubric (\S\ref{sec:exp}), not percentages: $5$ is a complete match in agent and reasoning, $0$ is irrelevant or wrong. The best fault ($2.27/5.0$) and violation ($0.82/5.0$) scores thus indicate \emph{partially} correct predictions---often naming a relevant agent but missing or misattributing the causal reasoning---rather than random outputs.

\paragraph{Human validation of LLM-as-Judge.}
On $N=45$ samples from Cosmos-Reason2-2B (Fine-tuned), two annotators applying the same rubric achieved moderate agreement ($\kappa=0.579$).
GPT-4o strongly correlates with mean human scores (Spearman $\rho=0.851$, Pearson $r=0.882$, $p<0.001$), supporting judge reliability; we treat the judge as a consistent ranking signal rather than an absolute oracle (Sec.~\ref{sec:limitations}).

\paragraph{Accident-type confusion analysis.}
Figure~\ref{fig:confusion} aggregates accident-type predictions over all six models. Both base and fine-tuned regimes collapse onto \emph{Rear-End} (over-predicted $2.58\times$ and $2.01\times$, respectively) and almost never recover \emph{Side-by-Side} ($\approx$1\% recall) or \emph{Head-on}, reflecting a strong class prior rather than scene understanding. Fine-tuning mainly redistributes mass toward \emph{None}, which raises overall accuracy modestly ($33.1\%\!\to\!35.4\%$) without correcting the underlying bias.

\subsection{Traffic-Rule Diversity and the Reasoning Bottleneck}
\label{sec:rulediv}

\paragraph{How diverse are the rules?}
A natural question is whether the low violation scores reflect a narrow rule space or genuine reasoning difficulty. We map every one of the $2{,}243$ free-text violation answers to a single \emph{primary} rule family using a deterministic, ordered keyword lexicon over an eleven-family ontology, where each family is defined by characteristic visual preconditions and typical evidence and a fixed priority resolves the $63\%$ of answers that cue multiple families (e.g., merging the former ``right-of-way at intersection'' into failure-to-yield). The mapping uses no LLM or human in the loop, so it is fully reproducible from the released lexicon; an author audit of 100 random answers found $95\%$ agreement with adjudicated labels, and non-answers or low-quality fragments ($0.3\%$ of test) are held out as ``unspecified'' rather than counted toward breadth. The space is broad (Table~\ref{tab:rule_diversity}): eleven distinct families (nine represented in the test split) led by unsafe-following-distance ($32.8\%$) and failure-to-yield/right-of-way ($30.6\%$), with a substantial tail of lane-change ($17.2\%$), turning/reversing, signal, speed, stopping, and pedestrian violations. Only $0.3\%$ of test answers are unspecified (non-answers such as ``N/A''), so coverage is not inflated by low-quality annotations. The benchmark therefore tests rule \emph{breadth}, not a single dominant rule. We also observe a train/test shift: loss-of-control and overtaking violations are present in CCD ($8.3\%$ and $1.3\%$ of train) but essentially absent from Nexar (test), reflecting differences in the source corpora.

\paragraph{Rule understanding vs.\ visual grounding.}
Per-family judge scores (Table~\ref{tab:rule_diversity}, averaged over all evaluated models) are uniformly low ($<\!1$ on the $0$--$5$ scale) but not uniform: families with clear relational cues (right-of-way, following-distance) fare best, whereas those whose visual signatures are subtle or easily confused (lane change, unsafe speed, sudden stop) are hardest. Inspecting predictions, the dominant error mode is not failing to know a rule but failing to ground it: models frequently cite a plausible rule yet attach it to the wrong agent or to an action not present in the video (hallucinated violations), and they confuse rules whose visual signatures overlap (e.g., following-distance vs.\ failure-to-yield at intersections). This is consistent with the gap lying in mapping visual evidence to the correct rule and agent, the same grounding limitation that LoRA on a frozen vision encoder does not address.

\begin{table}[t]
\centering
\caption{Traffic-rule violation families on CAViAR (train/test \% of violation answers) and per-family difficulty (mean LLM-as-Judge score, $0$--$5$, averaged over all evaluated models on the Nexar test split). Mapping uses a deterministic ordered lexicon over the ontology described in Sec.~\ref{sec:rulediv}; ``--'' denotes families with no test support. The space spans eleven families rather than a single dominant rule.}
\label{tab:rule_diversity}
\small
\setlength{\tabcolsep}{4pt}
\begin{tabular}{llccc}
\toprule
\textbf{Code} & \textbf{Violation family} & \textbf{Train\%} & \textbf{Test\%} & \textbf{Judge$\uparrow$} \\
\midrule
FD & Unsafe following distance / rear-end & 22.1 & 32.8 & 0.64 \\
RW & Failure to yield / right-of-way & 27.4 & 30.6 & 0.93 \\
LC & Improper lane change / merging & 12.8 & 17.2 & 0.33 \\
TU & Improper turn / U-turn / reversing & 6.9 & 5.9 & 0.59 \\
SG & Signal / sign violation & 5.5 & 5.5 & 0.60 \\
ST & Sudden stop / improper stopping & 1.3 & 2.3 & 0.08 \\
SP & Unsafe speed / reckless driving & 5.3 & 1.7 & 0.28 \\
AT & Inattentive / improper observation & 1.6 & 1.7 & 0.59 \\
PD & Pedestrian / non-motorized crossing & 0.9 & 0.3 & 0.96 \\
CT & Loss of vehicle control & 8.3 & 0.0 & -- \\
OT & Improper overtaking / passing & 1.3 & 0.0 & -- \\
\midrule
\multicolumn{2}{l}{Other (rare valid rules)} & 5.4 & 1.6 & 0.35 \\
\multicolumn{2}{l}{Unspecified / low-quality} & 1.2 & 0.3 & 0.21 \\
\bottomrule
\end{tabular}
\end{table}
\section{Discussion}
\label{sec:discussion}

\paragraph{Anatomy of the Perception--Reasoning Gap.}
CAViAR's headline finding is not simply that responsibility reasoning is hard, but \emph{where} it breaks down. Perceptual competence is itself uneven---lighting is nearly solved and road condition strong in balanced accuracy, whereas weather sits at or below its majority baseline (Table~\ref{tab:mcq_diagnostics})---yet even attributes models read well do not transfer: the same models collapse on at-fault and especially rule-violation identification (Tables~\ref{tab:main_results},~\ref{tab:judge_breakdown}). Since the same clip supplies both context and responsibility labels, image quality, frame sampling, or domain shift alone cannot explain the gap, as those would also depress perception. The pattern is instead consistent with a reasoning bottleneck: models recover salient attributes yet fail to convert visual evidence into a rule-grounded responsibility judgment.

\paragraph{Why rule violation identification is the hardest subtask.}
Violation identification chains two capabilities: recognizing the relevant rule from the broad space documented in Sec.~\ref{sec:rulediv}, and grounding it in the specific agents and actions observed. As the per-family analysis shows (Sec.~\ref{sec:rulediv}), errors are dominated by the \emph{mapping} step rather than the \emph{recall} step, which is why scaling and fine-tuning---which mostly sharpen perception and phrasing---leave violation the weakest subtask.

\paragraph{Why 8B fine-tuning gains are marginal.}
Fine-tuning yields large gains for 2B models but little for 8B models (Table~\ref{tab:main_results}). We attribute this to three compounding factors. First, 8B base models already saturate the perception MCQs, leaving little headroom on the metrics most responsive to domain adaptation. Second, LoRA adapts only the language model while the vision encoder and projector are frozen, so additional visual grounding---the actual bottleneck for responsibility reasoning---is not learned. Third, the residual errors are reasoning-mapping errors (above), which are not closed by absorbing the surface style of CCD captions. Together these indicate that closing the gap will require grounding-aware training signals, not merely larger backbones.

\paragraph{Domain shift as a confound.}
Because training uses CCD and testing uses Nexar, distributional
differences between the two corpora are a competing explanation for low
test performance. Table~\ref{tab:label_shift} quantifies these shifts
with Total Variation Distance (TVD) and Jensen--Shannon Divergence (JSD):
Accident Type exhibits the largest shift (TVD$\,{=}\,0.362$, majority
class flips from T-Bone to None), followed by Lighting
(TVD$\,{=}\,0.268$, Night triples from 10.6\% to 37.4\%) and Road
Condition (TVD$\,{=}\,0.192$, Snowy vanishes). The rule-violation family
distribution also shifts (TVD$\,{=}\,0.182$): loss-of-control and unsafe
speed are present in CCD but nearly absent from Nexar, while
following-distance and lane-change violations are over-represented in
test.
However, the CCD holdout results suggest that domain shift is not the only
factor behind the low responsibility scores, although it remains a meaningful
confound. Since both contextual and responsibility-oriented tasks are evaluated
on the same clips, the pattern of relatively stronger contextual performance
and consistently low responsibility scores is consistent with a bottleneck in
temporal grounding, agent-role assignment, and mapping visible actions to
annotated rule-relevant categories.

\paragraph{CCD holdout ablation.}
To isolate the contribution of domain shift, we construct a same-source
holdout split from CCD alone: 1{,}200 videos for training and 300 for
testing, so train and test share the same corpus distribution. We
evaluate Qwen3-VL (2B/8B, LoRA fine-tuned on the 1{,}200 CCD holdout
train) and Cosmos-Reason2 (2B/8B, zero-shot base) on the 300-video CCD
holdout test. Table~\ref{tab:holdout} reports the results. MCQ accuracy
ranges from 62.17\% to 67.92\%, comparable to the cross-corpus Nexar
results (Table~\ref{tab:main_results}), and BERTScore-F1 ranges from
31.12 to 39.60---again in the same band as the main experiments. If
domain shift were the primary bottleneck, same-source evaluation should
yield substantially higher scores; instead, the Perception--Reasoning
Gap persists even when the CCD$\to$Nexar distributional shift is
eliminated. This provides evidence that low responsibility-reasoning performance is not
solely an artifact of CCD-to-Nexar distribution shift.

\begin{table}[t]
\centering
\caption{CCD holdout ablation (same-source split: 1{,}200
train / 300 test, all CCD). Per-task MCQ accuracy (\%) and overall
BERTScore-F1 (\%) for open-ended tasks. Reasoning scores remain low
even without the CCD$\to$Nexar domain shift.}
\label{tab:holdout}
\small
\setlength{\tabcolsep}{3pt}

\begin{tabular}{lcccccc}
\toprule
\textbf{Model} & \textbf{Setting} & \textbf{Weather} & \textbf{Acc.\ Type} & \textbf{Road} & \textbf{MCQ Overall} & \textbf{BERT-F1} \\
\midrule
Qwen3-VL-2B & FT & 84.67 & 12.67 & 77.67 & 64.92 & 39.60 \\
Qwen3-VL-8B & FT & 77.33 & 34.33 & 82.67 & 67.92 & 33.66 \\
Cosmos2-2B   & Base & 77.67 & 31.33 & 62.00 & 62.17 & 34.46 \\
Cosmos2-8B   & Base & 74.00 & 35.33 & 71.67 & 63.75 & 31.12 \\
\bottomrule
\end{tabular}
\end{table}

\begin{table}[t]
\centering
\caption{Train/test label shift between CCD (train) and Nexar (test).
TVD = Total Variation Distance (0 = identical, 1 = disjoint);
JSD = Jensen--Shannon Divergence (bits). Accident Type shows the largest
shift; Weather is relatively stable.}
\label{tab:label_shift}
\small
\setlength{\tabcolsep}{4pt}
\begin{tabular}{lccl}
\toprule
\textbf{Task / Distribution} & \textbf{TVD}$\uparrow$ & \textbf{JSD (bits)}$\uparrow$ & \textbf{Key shift} \\
\midrule
Weather        & 0.104 & 0.020 & Snowy vanishes (2.3\%$\to$0\%) \\
Lighting       & 0.268 & 0.074 & Night triples (10.6\%$\to$37.4\%) \\
Road Condition & 0.192 & 0.072 & Snowy vanishes (12.1\%$\to$0.1\%) \\
Accident Type  & 0.362 & 0.161 & Majority flips (T-Bone$\to$None) \\
\midrule
Rule-Viol.\ Family & 0.182 & 0.063 & Control/speed vanish; FD rises \\
\bottomrule
\end{tabular}
\end{table}

\subsection{Limitations and Future Work}
\label{sec:limitations}

Several limitations frame our claims and point to future work. (i)~\emph{Annotation reliability:} CAViAR uses a primary-annotation plus quality-control review protocol (two annotators, two reviewers) but we do not yet report formal inter-annotator agreement; a stratified independent re-annotation study is needed. (ii)~\emph{Evaluation metrics:} BERTScore and LLM-as-Judge are imperfect proxies for agent-role and rule-relevant responsibility content; we mitigate this by reporting MCQ accuracy, BERTScore-F1, and GPT-4o judging jointly and validating the judge against humans (Sec.~\ref{sec:exp}), but a larger human--judge study beyond our $45$-sample validation is left to future work. (iii)~\emph{Human baseline:} we do not yet report lay or expert human performance, which would calibrate the difficulty ceiling. (iv)~\emph{Model coverage:} we focus on open-weight VLMs for reproducibility; large proprietary and reasoning-tuned systems would map the current frontier. (v)~\emph{Bias attribution:} we document an accident-type prediction bias (Fig.~\ref{fig:confusion}) but do not yet attribute it to specific visual evidence via saliency analysis.
\section{Conclusion}

We introduced CAViAR, a real-world dashcam benchmark with 2,249 videos and 20,108 QA pairs spanning perception, environmental conditions, causal explanation, and responsibility attribution. By adding structured supervision for apparent at-fault agent, affected agent, and apparent rule-violation category, CAViAR enables systematic evaluation of safety-critical reasoning beyond accident detection or generic VideoQA. Our experiments expose a consistent Perception--Reasoning Gap: state-of-the-art VLMs handle salient perceptual cues unevenly yet struggle far more to convert visual evidence into annotated rule-relevant responsibility categories, with violation identification the hardest subtask even after fine-tuning. We hope CAViAR catalyzes work on grounding-aware multimodal reasoning for autonomous safety.

\section{Dataset Release, Ethics, and Misuse Safeguards}
\label{sec:ethics}

\textbf{Release and licensing.}
We will release CAViAR annotations, train/test splits, evaluation scripts, prompt templates, frame indices, and reproduction metadata. Original videos remain governed by the CCD and Nexar licenses; where redistribution is not permitted we release identifiers, links, and derived annotations rather than video files. CAViAR is intended only for academic research on accident understanding and rule-relevant multimodal reasoning.

\textbf{No legal or punitive use.}
CAViAR labels are video-grounded research annotations of \emph{apparent} responsibility cues, not legal determinations of liability, negligence, or traffic-law violation. They must not be used for legal adjudication, insurance, employment screening, policing, enforcement, surveillance, driver profiling, or any decision about an identifiable individual, vehicle, or organization.

\textbf{Privacy, consent, and bias.}
CAViAR is built from existing public-road dashcam corpora, so privacy handling follows the source datasets; we add no new personal identifiers and, where raw video is redistributed, encourage blurring faces, plates, and GPS overlays. Explicit consent from all visible road users is generally impractical for such footage, further restricting use to aggregate model analysis. The benchmark inherits geographic, camera, fleet, and selection biases from its sources, reflects an ego-vehicle viewpoint that can miss off-screen or occluded causes, and may overrepresent common patterns (e.g., rear-end) while underrepresenting vulnerable road users---limiting the generality of conclusions. We will provide a public contact channel for privacy concerns, annotation errors, and takedown requests, with validated changes documented in a versioned changelog.




{
    \small
    \bibliographystyle{ieeenat_fullname}
    \bibliography{main}
}
\end{document}